\documentclass[journal]{IEEEtran}

\ifCLASSINFOpdf
\else
   \usepackage[dvips]{graphicx}
\fi
\usepackage{url}
\usepackage{graphicx}
\usepackage{multirow}
\usepackage{enumerate}
\usepackage{amsmath,amsfonts}
\usepackage{algorithmic}
\usepackage{algorithm}
\usepackage{array}
\usepackage{longtable}
\usepackage{booktabs}
\usepackage{adjustbox}
\usepackage{authblk}
\usepackage{cite}
\usepackage{balance}
\usepackage{arydshln}
\begin{document}

\title{AI-Generated Video Detection via Spatio-Temporal Anomaly Learning}

\author{Jianfa Bai$^{\dag}$, Man Lin$^{\dag}$, and Gang Cao$^{\ast}$,  \IEEEmembership{Member, IEEE}
\thanks{$^{\dag}$Contributed equally. $^{\ast}$Corresponding author, e-mail: gangcao@cuc.edu.cn.}
% \thanks{This paragraph of the first footnote will contain the date on which you submitted your paper for review. It will also contain support information, including sponsor and financial support acknowledgment. For example, ``This work was supported in part by the U.S. Department of Commerce under Grant BS123456.'' }
\thanks{J. Bai, M. Lin and G. Cao are with the State Key Laboratory of Media Convergence and Communication, Communication University of China, and also with the School of Computer and Cyber Science, Communication University of China, Beijing 100024, China.}}
% \thanks{S. B. Author, Jr., was with Rice University, Houston, TX 77005 USA. He is now with the Department of Physics, Colorado State University, Fort Collins, CO 80523 USA (e-mail: author@lamar.colostate.edu).}}

% \markboth{Journal of \LaTeX\ Class Files, Vol. 14, No. 8, August 2015}
% {Shell \MakeLowercase{\textit{et al.}}: Bare Demo of IEEEtran.cls for IEEE Journals}
 \maketitle

\begin{abstract}
The advancement of generation models has led to the emergence of highly realistic artificial intelligence(AI)-generated videos. Malicious users can easily create non-existent videos to spread false information. This letter proposes an effective AI-generated video detection (AIGVDet) scheme by capturing the forensic traces with a two-branch spatio-temporal convolutional neural network (CNN). Specifically, two ResNet sub-detectors are learned separately for identifying the anomalies in spatical and optical flow domains, respectively. Results of such sub-detectors are fused to further enhance the discrimination ability. A large-scale generated video dataset (GVD) is constructed as a benchmark for model training and evaluation. Extensive experimental results verify the high generalization and robustness of our AIGVDet scheme. Code and dataset will be available at https://github.com/multimediaFor/AIGVDet.
\end{abstract}

\begin{IEEEkeywords}
Video forensics, Generated video detection, Spatial-temporal anomaly, Optical flow, Decision fusion.
\end{IEEEkeywords}

\IEEEpeerreviewmaketitle

\section{Introduction}

\IEEEPARstart{T}{he} development of large models has greatly propelled the advancement of AI-generated content. AI-generated videos with exceptional quality, rapid creation and cost-effectiveness are revolutionizing industries, such as short and long-form video production, gaming and advertising. However, there also appear high risks associated with such videos including the spread of misinformation and manipulation of public opinion. Many generated videos are so realistic that they are virtually indistinguishable from real ones, particularly with the latest generation models like Sora \cite{sora}. Despite regulatory attempts such as Biden's signing of AI act \cite{bidensign}, reliable blind detection tools are still necessary to differentiate between generated and real videos.

As a new digital forensics problem, there are none prior works on the blind detection of detecting AI-generated videos. Many forensic methods \cite{wang2020cnn, gragnaniello2021gan, corvi2023detection, wang2023dire} have been proposed for detecting AI-generated images. In \cite{wang2020cnn, gragnaniello2021gan}, large-scale training dataset and data augmentation strategies are employed to learn common systematic flaws in GAN generated images, resulting in detectors with good generalization. However, as found in our experiments, the performance of such methods becomes poor in detecting generated videos. That should attribute to the different generation mechanisms between synthesized images and videos, such as spatial-temporal consistency. There are also many prior works \cite{caldelli2021optical,yang2020preventing,gu2022delving} in detecting forged face videos. Yang et al. \cite{yang2020preventing} propose to exploit the appearance and motion characteristics of lips to defend against sophisticated DeepFake attacks. Gu et al. \cite{gu2022delving} design different modules to capture the inconsistencies caused by subtle movements within and between densely sampled video snippets. Such methods focus on capturing physiological defects and subtle facial abnormalities in manipulated videos, which are not suitable for detecting generated videos with various scenes. 

\begin{figure}[!t]
  \centering
  \includegraphics[width=\linewidth]{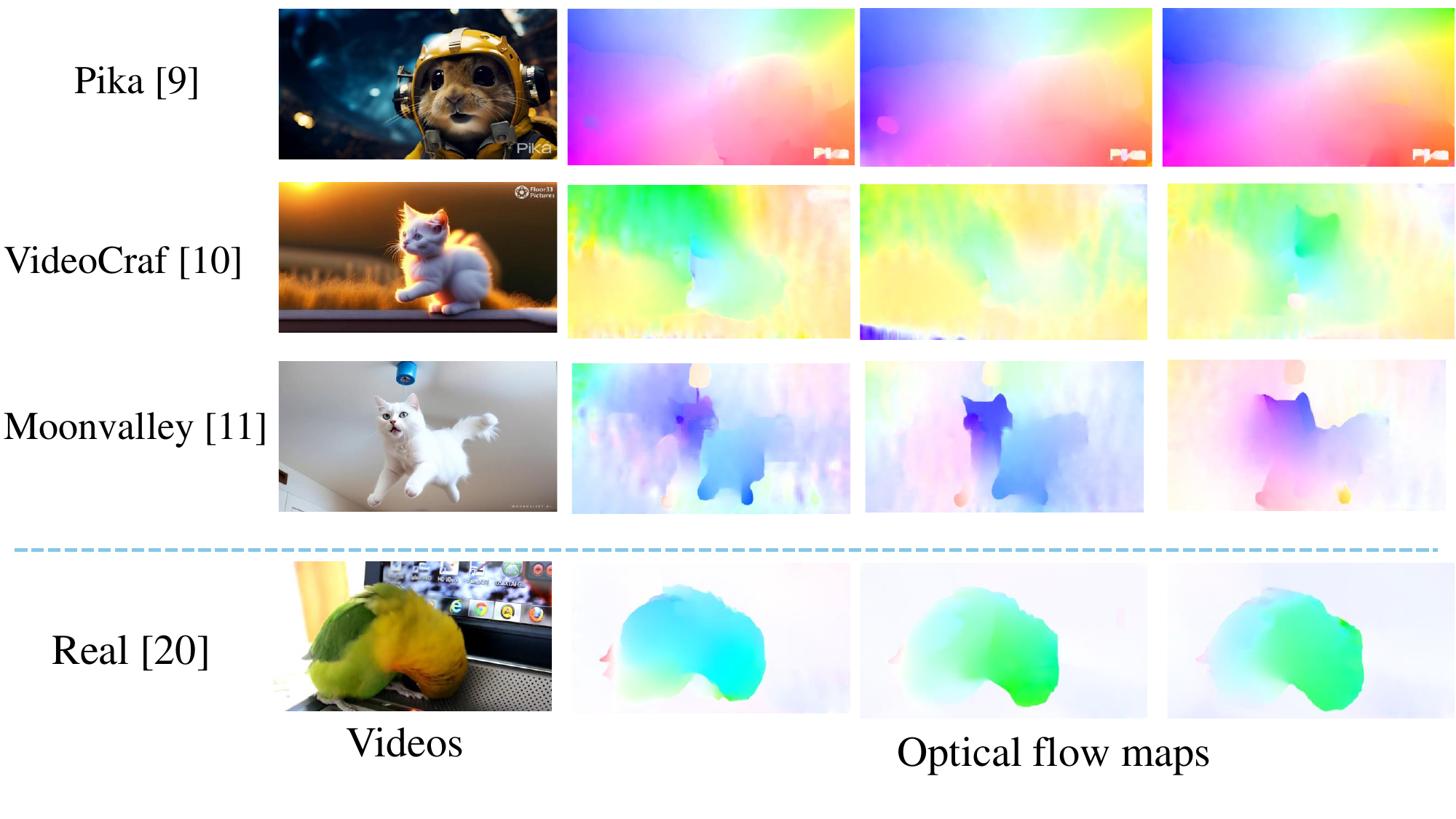}
  \vspace{-20pt}
  \caption{The first frame and the first three optical flow maps of the videos generated by three models, along with those of a real video. }
  \label{fig}
\end{figure}

\begin{figure*}[!ht]
  \centering
  \includegraphics[width=\textwidth]{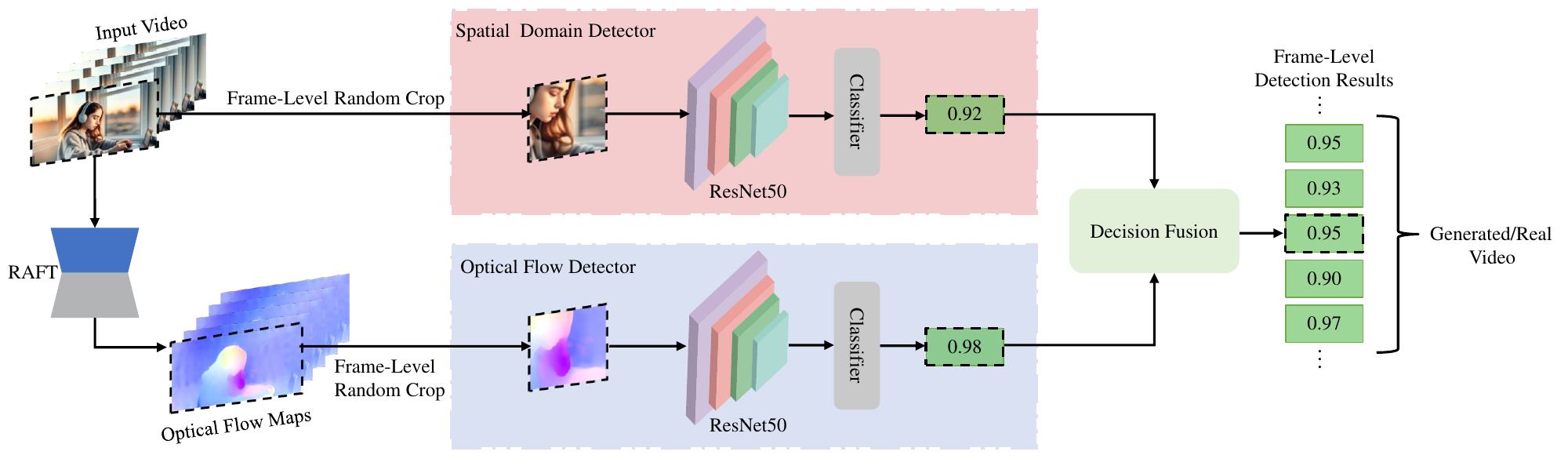}
  \caption{Overall pipeline of the proposed generated video detection scheme AIGVDet, where RAFT \cite{teed2020raft} is the method for calculating optical flow maps.}
\end{figure*}

To address such a gap in existing works, we formally point out the AI-generated video detection problem, and propose an effective solution. We observe that low-quality generated videos may exhibit some anomalies, such as abnormal textures and violations of physical rules in the frames. High-quality generated videos, which are indistinguishable from real ones to the naked eye, are likely to manifest temporal discontinuities in optical flow maps. Fig. 1 illustrates some video frames and optical flow maps estimated with the RAFT \cite{teed2020raft} method, where color indicates movement direction and shade represents movement magnitude. Despite the remarkable visual fidelity of generated video frames, their optical flow maps exhibit less smoothness and blurry contours compared to real ones. To capture such disparities between real and generated samples, we propose a simple yet effective AI-generated video detection (AIGVDet) model. The RGB frames and their corresponding optical flow maps serve as inputs, with a two-branch ResNet50 \cite{he2016resnet} encoder thoroughly exploring abnormalities in both modalities. In the end, a decision-level fusion binary classifier is assembled to effectively integrate information and enhance the model's discriminative capability. Additionally, we create a large-scale generated video benchmark dataset for network training and evaluation, comprising synthetic videos from 11 different generator models. 
% Extensive experiments are conducted to analyze factors contributing to cross-model generalization. 
Extensive experiments showcase the generalization capability and robustness of our proposed detector.

The rest of this letter is organized as follows. The generated video detection scheme is proposed in Section II, followed by extensive experiments and discussions in Section III. The conclusions are drawn in Section IV.

\section{Proposed AIGVDet Scheme}
This section begins by presenting our proposed generated video detector AIGVDet. Then the newly constructed generated video dataset GVD is introduced in detail.

\subsection{Two-branch Spatio-temporal Detector: AIGVDet}
The overall pipeline of our AIGVDet scheme is illustrated in Fig. 2. It comprises two individually trained detectors both with ResNet50 \cite{he2016resnet} backbone network, namely the spatial domain and the optical flow detectors. The former explores the abnormality of spatial pixel distributions (e.g. texture, noise, etc.) within single RGB frames, while the latter captures temporal inconsistencies via optical flow. Let a video be denoted by $N$ frames $\{{{I}}_{i}\}_{i=1}^{N}\in\mathbb{R}^{W \times H \times 3}$ with a spatial resolution of $W \times H$ pixels. Then the optical flow maps $F_{i}\in\mathbb{R}^{W \times H \times 3}$ between adjacent frames are calculated by the RAFT \cite{teed2020raft} estimator $ \mathcal{F}(\cdot)$ as
\begin{align}
&F_{i}={\mathcal{F}}(\,I_{i},\;I_{i+1}). \tag{1}
\end{align}
Each frame ${I}_{i}$ and its corresponding optical flow ${F}_{i}$ are then randomly cropped to 448x448, and fed into the spatical domain encoder $\mathrm{R}(\cdot)$ and the optical flow branch encoder $\mathrm{O}(\cdot)$, respectively. Subsequently, the extracted two-modal features $V_{I}^{i}$ and $V_{F}^{i}$ are sent to two binary classifiers, both consist of a global average pooling layer $\mathrm{GAP}(\cdot)$, a fully connected layer $\mathrm{FC}(\cdot)$ and a sigmoid activation function $\mathrm{Sigmoid}(\cdot)$. This process maps the two feature vectors to probabilities ${P}_{I}^{i}$ and ${P}_{F}^{i}$, indicating the likelihood of being generated. That is,
\begin{align}
&{V}_{I}^{i}={\mathrm{R}}({I}_{i}), \quad {V}_{F}^{i}={\mathrm{O}}({F}_{i}).  \tag{2}\\
&{P}_{I}^{i}=\mathrm{Sigmoid}(\mathrm{FC}(\mathrm{GAP}(V_{I}^{i}))), \nonumber\\
&{P}_{F}^{i}=\mathrm{Sigmoid}(\mathrm{FC}(\mathrm{GAP}(V_{F}^{i}))). \tag{3}
\end{align}

To integrate the representational capabilities of these two feature vectors, we employ decision-level  fusion to obtain the prediction result ${P}^{i}$ for each frame. It is the weighted summation of ${P}_{I}^{i}$ and ${P}_{F}^{i}$ as
\begin{align}
&P^{i}=\alpha {{P}_{I}^{i}}+\left(1-\alpha\right){{P}_{F}^{i}}, \tag{4}
\end{align}
where $\alpha\in\left(0,1\right)$ is a weight to balance the predictions from spatial domain and optical flow modalities. Lastly, the final video-level prediction  result ${P}$ is computed as
\begin{align}
&P={\frac{1}{N-1}\sum}_{i=1}^{N-1}{P}^{i}. \tag{5}
\end{align}
Here, $P\in\left(0,1\right)$ signifies the probability for being a AI-generated video. 

Note that the state-of-the-art video generation algorithms still encounter some technical bottlenecks on realistic generation. Due to a lack of deep understanding of physical laws, the generated content may deviate from real-world physics. There may exist discrepancies in the direction and speed of object motion compared to reality. Such findings  motivate us to leverage optical flow maps to enhance the discrimination of generated videos. The optical flow is an estimation of the per-pixel movement between neighboring frames. We employ the RAFT \cite{teed2020raft} algorithm to compute the optical flow prediction between adjacent frames. It comprises three main components, i.e., feature extraction, computing visual similarity, and iterative updates \cite{teed2020raft}.
% \subsubsection{Feature Extraction}
% Given two adjacent frames ${I}_{1}$ and ${I}_{2}$, the feature enconder network $g_\theta$ is applied to them to map the input frames to a density feature map of lower resolution, while the context network $h_\theta$ extracts features only from ${I}_{1}$.

% \subsubsection{Computing Visual Similarity}
% After extracting features $g_\theta\left(I_1\right)$ and $g_\theta\left(I_2\right)$,the correlation volume $\mathbf{C}$ is formed by calculating the dot product between all pairs of feature vectors product.

% \subsubsection{Iterative Updates}
% The update factor estimates a series of flow estimates$\left\{{f}_1, \ldots, {f}_N\right\}$ starting from ${f}_0=0$. With each iteration, it produces a new update direction $\Delta {f}$, which is applied to estimating the current estimate ${f}_{k+1}=\Delta {f}+{f}_{k+1}$. During the update process, the core component of the update operator is the gated activation unit based on the GRU unit, in which the fully connected layers are replaced by convolutions. The resolution of the output flow ${F}$ is 1/8 of the input frame, and is finally upsampled to match the actual resolution.

% \noindent\textbf{ Decision-level feature fusion. }
% \vspace{-3pt}
\subsection{Generated Video Dataset}

% \begin{table}[h]
% \begin{table}[!tb]
% \caption{Details of our collected generated video dataset (GVD).}
% \centering
% \begin{adjustbox}{width=\linewidth}
% \setlength{\tabcolsep}{3pt}
% \begin{tabular} {clllll}
% % \begin{tabular} {clccccc}
% % {p{2.cm}p{4.0cm}p{2cm}}
% % {p{2.cm}p{4.0cm}p{2cm}p{2cm}p{2cm}}
% \hline
% % \textbf{Family/Method} & \textbf{Video Source} & \textbf{\# Videos} \\
% \textbf{Type} & \textbf{Name} & \textbf{Number}& \textbf{Resolution}& \textbf{Format}& \textbf{Source} \\
% \hline
% \multirow{7}{*}{T2V} 
% & Moonvalley \cite{Moonvalley} & 3.55k & 1184*672 & MP4  & Discord  \\
% & VideoCraft \cite{chen2023videocrafter1} & 1.5k & 1024*576 & MP4 & Discord \\
% & Pika \cite{pika} &  1.0k & Variable  & MP4 & Discord \\
% & NeverEnds \cite{neverends} &  1.0k  &  1024*576 & MP4 &  Discord \\
% & Emu \cite{girdhar2023emu} &  900 &  512*512 & MP4 & Discord \\
% & VideoPoet \cite{kondratyuk2023videopoet} &  120 & 512*896 & MP4 & Official Web.\\
% & Hotshot \cite{hotshot} &  500 &  672*384 & GIF & Official Web.\\
% & Sora \cite{sora} & 48 & Variable & MP4 & Official Web.\\
% \hdashline % 在表格中绘制水平虚线
% \multirow{3}{*}{I2V} 
% & Moonvalley \cite{Moonvalley} & 1.0k & Variable  & MP4 & Discord \\
% & Pika \cite{pika} &  1.0k & Variable  & MP4 & Discord \\
% & NeverEnds \cite{neverends} &  1.0k&  Variable & MP4 & Discord \\
% \hline
% \end{tabular}
% \end{adjustbox}
% % \caption{Video Source and Number of videos for Different Family/Methods}
% \end{table}
% % \vspace{-20pt}

To train and evaluate the AIGVDet model, the GVD is constructed by collecting 11,618 video samples yielded by 11 state-of-the-art generator models. Each generator model is trained on a distinct real video dataset tailored to its specific task. The most two common types of generation models, i.e., Text-to-Video (T2V) and  Image-to-Video (I2V) are involved specifically. T2V refers to automatic generation of corresponding videos based on the content described in the text. I2V refers to generating videos using the images accompanied by descriptive information, or just images. Specific details of the GVD are presented in Table I. The main collection source ‘Discord’ \cite{Discord} is a free network communication and digital distribution platform. Within the video-generating clubs of such a platform, users can share and showcase their videos synthesized using various generator models.

\begin{table}[!tb]
\caption{Details of our collected generated video dataset (GVD).}
\centering
\begin{adjustbox}{width=\linewidth}
\setlength{\tabcolsep}{3pt}
\begin{tabular} {clllll}
% \begin{tabular} {clccccc}
% {p{2.cm}p{4.0cm}p{2cm}}
% {p{2.cm}p{4.0cm}p{2cm}p{2cm}p{2cm}}
\hline
% \textbf{Family/Method} & \textbf{Video Source} & \textbf{\# Videos} \\
\textbf{Type} & \textbf{Name} & \textbf{Number}& \textbf{Resolution}& \textbf{Format}& \textbf{Source} \\
\hline
\multirow{7}{*}{T2V} 
& Moonvalley \cite{Moonvalley} & 3.55k & 1184*672 & MP4  & Discord  \\
& VideoCraft \cite{chen2023videocrafter1} & 1.5k & 1024*576 & MP4 & Discord \\
& Pika \cite{pika} &  1.0k & Variable  & MP4 & Discord \\
& NeverEnds \cite{neverends} &  1.0k  &  1024*576 & MP4 &  Discord \\
& Emu \cite{girdhar2023emu} &  900 &  512*512 & MP4 & Discord \\
& VideoPoet \cite{kondratyuk2023videopoet} &  120 & 512*896 & MP4 & Official Web\\
& Hotshot \cite{hotshot} &  500 &  672*384 & GIF & Official Web\\
& Sora \cite{sora} & 48 & Variable & MP4 & Official Web\\
\hdashline % 在表格中绘制水平虚线
\multirow{3}{*}{I2V} 
& Moonvalley \cite{Moonvalley} & 1.0k & Variable  & MP4 & Discord \\
& Pika \cite{pika} &  1.0k & Variable  & MP4 & Discord \\
& NeverEnds \cite{neverends} &  1.0k&  Variable & MP4 & Discord \\
\hline
\end{tabular}
\end{adjustbox}
% \caption{Video Source and Number of videos for Different Family/Methods}
\end{table}
% \vspace{-20pt}

% \textbf{Authentic Video}. For the authentic video in each subset, we randomly take an equal number of authentic videos from the GOT dataset[6] with the generated videos, which are not from the same source as the training and validation sets in our experiments.

% \subsection{The method to train}
% In our training, the video is first divided into frames, then the optical flow between adjacent frames is calculated by RAFT[5]. Additionally, the optical flow is randomly cropped with the size of 448*448 and horizontally flipped with a probability of 0.5. Each optical flow is put into a Resnet50 [7] network that is pre-trained with ImageNet as a unit, yielding a number between 0 and 1. If this number exceeds the threshold we set (0.1 in this experiment), the optical flow is considered generated video's optical flow; otherwise, it is classified as real video's optical flow.
% and then passed through a fully connected layer and sigmoid layer to determine whether it is generated, 
% see Fig. 2.

\section{Experiments}

\subsection{Experimental Setup}
The spatial domain detector and the optical flow detector are trained separately. All training steps are conducted in the same way apart from the difference in input. Further details regarding datasets, preprocessing methods, and evaluation metrics are elaborated below.

\noindent\textbf{Datasets.}
To comprehensively explore the generalization of our AIGVDet, we conduct training solely on generated videos from a single generation model, while testing on videos from various sources to simulate real-world scenarios. Specifically, we utilize 550 T2V generated videos from the Moonvalley \cite{Moonvalley} and 550 real videos from the YouTube\_vos2 dataset \cite{yang2019video} for model training and validation, with a training-validation set ratio of 10:1. Each video is sampled 95 RGB frames and their corresponding 94 optical flow maps. 
% Hence, for each detector, there are a total of 94,000 images for training and 9,400 images for validation. 
All generated videos in GVD, except for those used for model training and validation, will be utilized for testing. The corresponding number of real test videos is sourced from the GOT dataset\cite{huang2019got}.

\noindent\textbf{Preprocessing.} Since some collected generated videos contain watermarks at the bottom, the frames are firstly cropped to remove such watermark for avoiding bias. During training, considering that the frames of real videos in the training dataset are all in JPEG format, we randomly compress the generated frames with a JPEG compression factor ranging from 70 to 90 to mitigate the impact of compression. Prior to being fed into the detectors, all input images are cropped to a size of 448x448. Random cropping is employed during training, while central cropping is utilized during testing.

\noindent\textbf{Comparative methods and metrics.} Comparable detectors specifically designed for generated videos are lacking, thus we compare against a pioneering and classic detector for generated images, namely Wang \cite{wang2020cnn}. To ensure a fair comparison, we retrain the Wang using its provided source code on the same dataset utilized in our study. The training and testing strategies are consistent with its original settings, where RGB images serve as input, resized to 256x256 and then cropped to 224x224. We use accuracy (ACC) and area under the receiver-operating curve (AUC) as the video-level evaluation metrics. The video-level classification threshold is set at 0.1 and the weight $\alpha$ during decision-level fusion is set at 0.5.
% The third reason is that the both datasets' frame rates are similar.

\noindent\textbf{Implementation.} The proposed method is implemented using the PyTorch deep learning framework, and all experiments are conducted on an A800 GPU. Binary cross-entropy is used as the loss function. The backbone ResNet50 \cite{he2016resnet} is pre-trained with ImageNet \cite{deng2009imagenet}. We use the Adam optimizer as optimization function with an initial learning rate 1e-4. The learning rate is reduced by a factor of 10 if the validation accuracy does not increase after 5 epochs. Training is terminated when the learning rate reaches 1e-6. Data augmentation is applied to 10\% of training samples, includes Gaussian blurring with a sigma of 0.5, JPEG compression with a compression factor of 75, and random flipping. 

\subsection{Ablation Study}
% \begin{table}[h]
\begin{table}[tb]
\caption{Ablation test results. ACC (\%) and AUC (\%) among different variants of proposed scheme on T2V generated videos. 
% A refers to the original RGB image input, B refers to the optical flow input without random cropping, and C refers to the optical flow input without data augmentation. D refers to the optical flow input after random cropping and data augmentation. 
 }
\centering
% \begin{tabular}{p{4.0cm}<{\centering}p{5.0cm}}
\begin{adjustbox}{width=\linewidth}
\setlength{\tabcolsep}{1pt}
\begin{tabular}{lccccc}
% \begin{tabularx}{\linewidth}{lccccc}
\toprule
% \hline
\textbf{Variants}   & Moonvalley & VideoCraft & Pika & NeverEnds & Average\\ \midrule
% \hline
$\mathrm{S}_{spatial}$   &  96.8/\textbf{100} & 85.1/93.9 & 83.7/93.7 & 80.4/92.0 &  86.5/94.9\\
$\mathrm{S}_{optical}$   &  93.4/99.8 & 89.1/95.0 & 86.6/92.7 & 86.8/92.6   & 89.0/95.0  \\
$\mathrm{S}_{optical\_no\_cp}$  & 95.4/99.8 &  88.7/94.4 & 80.3/89.8 & 85.3/91.7  & 87.4/93.9 \\
$\mathrm{FF}_{concat}$   &  \textbf{99.5}/\textbf{100} & 73.0/92.2 & 69.7/92.2 & 75.1/94.1 & 79.3/94.6\\
$\mathrm{FF}_{add}$   &  97.2/\textbf{100} & 56.4/90.4 & 53.4/87.6 & 54.7/89.4 &  65.4/91.9\\
AIGVDet &  95.1/\textbf{100} & \textbf{91.3}/\textbf{97.0} & \textbf{89.5}/\textbf{95.5} & \textbf{89.5}/\textbf{95.7} & 
 \textbf{91.4}/\textbf{97.1} \\
% Average &  83.8/94.5 & 88.3/94.6 & 87.3/94.7 & 89.3/95.9 & \textbf{90.0}/\textbf{96.0} \\
% RGB-Optical Flow & 0.71 & 0.61 & 7 & 5 & 9\\
\bottomrule
% \hline
\end{tabular}
\end{adjustbox}
% \end{tabularx}
\end{table}
% \vspace{-10pt}

We conduct multiple ablation experiments to analyze the factors contributing to the cross-model generalization of this scheme. For this purpose, we train multiple variants using the same settings as stated above and assess their performance on the subsets of T2V generated videos. These variants include: trained solely on RGB frames ($\mathrm{S}_{spatial}$), trained solely on optical flow maps ($\mathrm{S}_{optical}$), trained solely on optical flow maps without cropping operation ($\mathrm{S}_{optical\_no\_cp}$), trained simultaneously on RGB frames and optical flow maps with feature fusion through concatenation ($\mathrm{FF}_{concat}$), trained simultaneously on RGB frames and optical flow maps with feature fusion through element-wise addition ($\mathrm{FF}_{add}$). The results of comparative experiments are shown in Table II. 

\noindent\textbf{Efficacy of feature representation.} Comparing the results of $\mathrm{S}_{spatial}$ and $\mathrm{S}_{optical}$ indicates that optical flow contain more discernible abnormal information than RGB frames, further illustrating that generated videos exhibit more instability in the temporal domain.

\noindent\textbf{Impact of preprocessing methods.} When directly training and testing the detector with entire optical flow maps without cropping,  $\mathrm{S}_{optical\_no\_cp}$ shows a 1.6\% decrease in average detection accuracy compared to $\mathrm{S}_{optical}$. The cropping operation enhances the detector's attention to local subtle changes in the optical flow maps while reducing interference from global information. This suggests that local information is more suitable for detecting generated videos than global information.

\noindent\textbf{Impact of fusion methods.} The RGB frames carry more visual details than optical flow maps, which also compensate for temporal information not contained in the RGB frames. Therefore, we study how to better fuse such two types of information. The results reveal that the decision fusion adopted by AIGVDet is more effective than the feature-level fusions via concatenation $\mathrm{FF}_{concat}$ and direct addition $\mathrm{FF}_{add}$. The detection network schemes based on feature fusion  tend to overfit on Moonvalley sub-dataset of T2V dataset and perform badly on the other ones. 

\subsection{Assessment of Generalization Ability}
\begin{table}[tb]
\caption{ ACC (\%) and AUC (\%) comparison of different detectors  on different test datasets. }
\centering
% \begin{tabular}{p{1.5cm}p{2.0cm}p{1.8cm}p{2.0cm}}
\begin{adjustbox}{width=\linewidth}
\setlength{\tabcolsep}{13pt}
\begin{tabular}{llcc}
\hline
\multicolumn{2}{c}{\textbf{Datasets}} & \multicolumn{2}{c}{\textbf{Detectors}} \\
\hline
\textbf{Type} & \textbf{Name} & \textbf{Wang\cite{wang2020cnn}} & \textbf{AIGVDet} \\
\hline
\multirow{7}{*}{T2V} 
& Moonvalley \cite{Moonvalley} & \textbf{99.3}/\textbf{100} & 95.1/\textbf{100} \\
& VideoCraft \cite{chen2023videocrafter1} & 75.7/92.3 & \textbf{91.3}/\textbf{97.0} \\
& Pika \cite{pika} & 68.9/89.5 & \textbf{89.5}/\textbf{95.5} \\
& NeverEnds \cite{neverends} & 81.7/95.5 & \textbf{89.5}/\textbf{95.7} \\
& Emu \cite{girdhar2023emu} & \textbf{97.3}/\textbf{99.7} & 94.1/99.3 \\
& VideoPoet \cite{kondratyuk2023videopoet} & 76.7/95.6 & \textbf{93.8}/\textbf{97.7} \\
& Hotshot \cite{hotshot} & 56.0/81.5 & \textbf{93.3}/\textbf{98.0} \\
& Sora \cite{sora} & 55.2/89.0 & \textbf{82.3}/\textbf{92.1}\\
\hdashline % 在表格中绘制水平虚线
&Average & 76.4/92.9 & \textbf{91.1}/\textbf{96.9} \\
\hline
\multirow{3}{*}{I2V} 
& Moonvalley \cite{Moonvalley} & 82.9/94.3 &\textbf{89.0}/\textbf{95.5} \\
& Pika \cite{pika} & 81.7/93.8 & \textbf{89.2}/\textbf{96.1} \\
& NeverEnds \cite{neverends} & 74.8/92.5 &\textbf{91.0}/\textbf{96.6} \\ 
\hdashline
&Average & 79.8/93.5  &\textbf{89.7}/\textbf{96.1}\\
\hline
\end{tabular}
\end{adjustbox}
% \caption{ Cross-generator generalization results. We report ACC (\%) and AUC (\%) (ACC/AUC in the Table).}
\end{table}
% \vspace{-10pt}

The varieties and quantities of video generation models will continue to evolve and improve, thus the generalization performance of detectors is crucial in real-world scenarios. Table III shows the detection results of Wang \cite{wang2020cnn} and our AIGVDet scheme on various types of generated videos.

Firstly, for T2V generation models, the AIGVDet achieves an accuracy of 95.1\% and an AUC of 100\% on the Moonvalley \cite{Moonvalley} dataset, which is consistent with the training set. It also achieves good results on other unseen models, especially the latest Sora \cite{sora}, reaching 82.3\%. Even for generated videos of different resolutions such as Videocraft \cite{chen2023videocrafter1} (1024x576), Emu \cite{girdhar2023emu} (512x512), Hotshot \cite{hotshot} (672x384), and VideoPoet \cite{kondratyuk2023videopoet} (512x896), effective discrimination can also be achieved. In average, the detection accuracy of AIGVDet (91.1\%) is significantly higher than that of Wang (76.4\%), and achieves 4\% AUC increment.

Our AIGVDet scheme also performs well on the more challenging I2V type of generated videos. Comparing with T2V, the I2V videos exhibits less pronounced motion with only subtle local changes, such as the ripples on water surface. Moreover, I2V videos are primarily generated using real image inputs, which further incurs challenges for detection. However, the results in Table III show that our average ACC (89.7\%) and average AUC  (96.1\%) surpass Wang \cite{wang2020cnn} by 9.9\% and 2.6\%, respectively. Such results demonstrate that our detector can generalize well to different types and resolutions of unseen generated videos.

\subsection{Robustness Evaluation}
We evaluate the robustness against video compression which is the most common post-processing of videos in real-life scenarios. CRF is a parameter that controls the compression quality of H.264. We test compression factors with CRF = 0, 18, 23, and 28 (0 indicates no compression). 
% As for the frame rate, we select the original frame rate of Moonvalley at 16. 
% For real videos, as the data itself consists of individual frames, we convert them to mp4 before detection. In the case of generated videos, since they are already in mp4 format, we extract frames, convert them to mp4, and then proceed with detection.
For both generated and real videos, we apply video recompression. We conduct this experiment on T2V videos from Moonvalley \cite{Moonvalley}, VideoCraft \cite{chen2023videocrafter1}, Pika \cite{pika}, NeverEnds \cite{neverends}, Hotshot \cite{hotshot} and Emu \cite{girdhar2023emu}. The results in Fig. 3 show that the AUC and ACC decreases to some extent with the increase of compression degree. However, the ACC for each detected model are all above 80\% , and the AUC consistently exceed 88\%.

% \begin{figure}
%     \centering
%     \begin{subfigure}
%         \centering
%         \includegraphics[height=5cm]{img/robust_mpeg_auc.png}
%         \caption{\textbf{Robutness.}We demonstrate the impact of AUC on MPEG under given test-time perturbations.}
%         \label{fig:img1}
%     \end{subfigure}%
%     \begin{subfigure}
%         \centering
%         \includegraphics[height=5cm]{img/robust_mpeg_acc.png}
%         \caption{\textbf{Robutness.}We demonstrate the impact of ACC on MPEG under given test-time perturbations.}
%         \label{fig:img2}
%     \end{subfigure}
%     \label{fig:double_img}
% \end{figure}

\begin{figure}[tb]
  \centering
  \includegraphics[width=\linewidth]{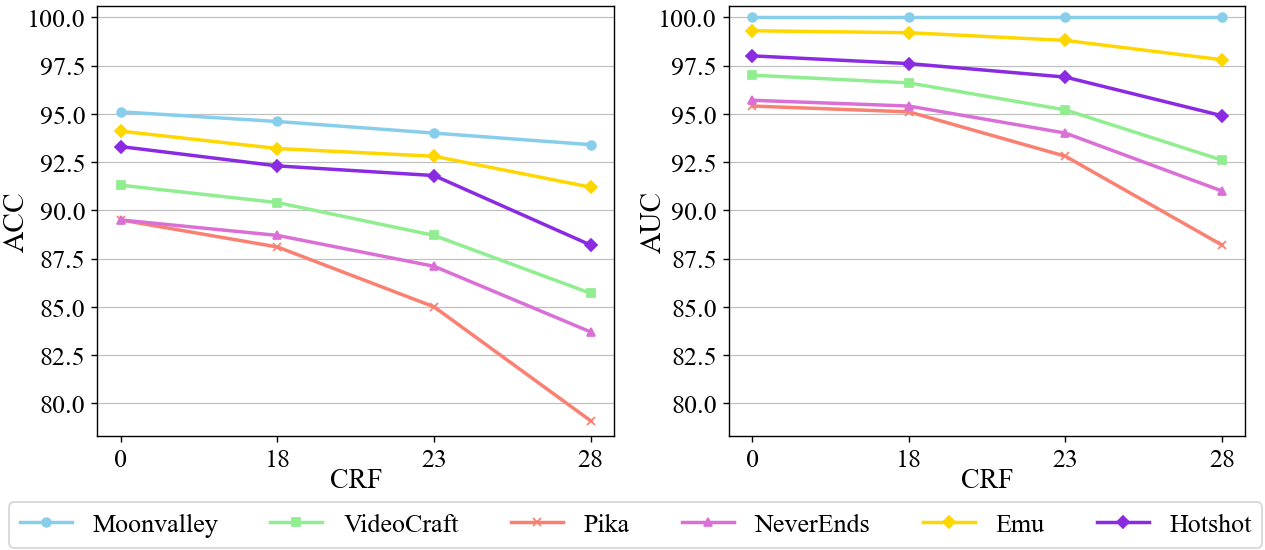}
  % \caption{\textbf{Robutness.}We demonstrate the impact of AUC on MPEG under given test-time perturbations.}
  \caption{Robustness evaluation results against post H.264 compression with different quality factors (CRFs).} 
  \label{fig}
\end{figure}

\section{Conclusion}
In this paper, we propose a simple yet effective scheme to detect AI-generated videos. Leveraging our newly constructed GVD, the proposed AIGVDet effectively capture and integrate spatial-temporal inconsistencies present in RGB frames and optical flow maps to distinguish between generated and authentic videos. Experimental results indicate that the state-of-the-art generated videos exhibit stronger instability in the temporal domain, which aids in discrimination. Moreover, our detection scheme exhibits good generalization to generated videos produced by various unknown generator models and is also effective against video compression. We aim for our work to serve as a robust baseline for detecting generated videos.

\bibliographystyle{IEEEtran}
\balance
\bibliography{references}

\begin{thebibliography}{10}
\providecommand{\url}[1]{#1}
\csname url@samestyle\endcsname
\providecommand{\newblock}{\relax}
\providecommand{\bibinfo}[2]{#2}
\providecommand{\BIBentrySTDinterwordspacing}{\spaceskip=0pt\relax}
\providecommand{\BIBentryALTinterwordstretchfactor}{4}
\providecommand{\BIBentryALTinterwordspacing}{\spaceskip=\fontdimen2\font plus
\BIBentryALTinterwordstretchfactor\fontdimen3\font minus \fontdimen4\font\relax}
\providecommand{\BIBforeignlanguage}[2]{{%
\expandafter\ifx\csname l@#1\endcsname\relax
\typeout{** WARNING: IEEEtran.bst: No hyphenation pattern has been}%
\typeout{** loaded for the language `#1'. Using the pattern for}%
\typeout{** the default language instead.}%
\else
\language=\csname l@#1\endcsname
\fi
#2}}
\providecommand{\BIBdecl}{\relax}
\BIBdecl

\bibitem{sora}
openai sora, \url{https://openai.com/sora}.

\bibitem{bidensign}
{The Washington Post}, \url{https://www.washingtonpost.com/technology/2023/10/30/biden-artificial-intelligence-executive-order/}.

\bibitem{wang2020cnn}
S.-Y. Wang, O.~Wang, R.~Zhang, A.~Owens, and A.~A. Efros, ``Cnn-generated images are surprisingly easy to spot... for now,'' in \emph{IEEE International Conference on Computer Vision and Pattern Recognition}, 2020, pp. 8695--8704.

\bibitem{gragnaniello2021gan}
D.~Gragnaniello, D.~Cozzolino, F.~Marra, G.~Poggi, and L.~Verdoliva, ``Are gan generated images easy to detect? a critical analysis of the state-of-the-art,'' in \emph{IEEE International Conference on Multimedia and Expo}, 2021, pp. 1--6.

\bibitem{corvi2023detection}
R.~Corvi, D.~Cozzolino, G.~Zingarini, G.~Poggi, K.~Nagano, and L.~Verdoliva, ``On the detection of synthetic images generated by diffusion models,'' in \emph{IEEE International Conference on Acoustics, Speech and Signal Processing}, 2023, pp. 1--5.

\bibitem{wang2023dire}
Z.~Wang, J.~Bao, W.~Zhou, W.~Wang, H.~Hu, H.~Chen, and H.~Li, ``Dire for diffusion-generated image detection,'' \emph{arXiv: 2303.09295}.

\bibitem{caldelli2021optical}
R.~Caldelli, L.~Galteri, I.~Amerini, and A.~Del~Bimbo, ``Optical flow based cnn for detection of unlearnt deepfake manipulations,'' \emph{Elsevier Pattern Recognition Letters}, vol. 146, pp. 31--37, 2021.

\bibitem{yang2020preventing}
C.-Z. Yang, J.~Ma, S.~Wang, and A.~W.-C. Liew, ``Preventing deepfake attacks on speaker authentication by dynamic lip movement analysis,'' \emph{IEEE Transactions on Information Forensics and Security}, vol.~16, pp. 1841--1854, 2020.

\bibitem{gu2022delving}
Z.~Gu, Y.~Chen, T.~Yao, S.~Ding, J.~Li, and L.~Ma, ``Delving into the local: Dynamic inconsistency learning for deepfake video detection,'' in \emph{AAAI Conference on Artificial Intelligence}, vol.~36, no.~1, 2022, pp. 744--752.

\bibitem{teed2020raft}
Z.~Teed and J.~Deng, ``Raft: Recurrent all-pairs field transforms for optical flow,'' in \emph{Springer European Conference on Computer Vision}, 2020, pp. 402--419.

\bibitem{he2016resnet}
K.~He, X.~Zhang, S.~Ren, and J.~Sun, ``Deep residual learning for image recognition,'' in \emph{IEEE International Conference on Computer Vision and Pattern Recognition}, 2016, pp. 770--778.

\bibitem{Discord}
Discord, \url{https://discord.com/}.

\bibitem{Moonvalley}
Moonvalley, \url{https://moonvalley.ai/}.

\bibitem{chen2023videocrafter1}
H.~Chen, M.~Xia, Y.~He, Y.~Zhang, X.~Cun, S.~Yang, J.~Xing, Y.~Liu, Q.~Chen, X.~Wang \emph{et~al.}, ``Videocrafter1: Open diffusion models for high-quality video generation,'' \emph{arXiv: 2310.19512}, 2023.

\bibitem{pika}
Pika, \url{https://www.pika.art/}.

\bibitem{neverends}
NeverEnds, \url{https://neverends.life}.

\bibitem{girdhar2023emu}
R.~Girdhar, M.~Singh, A.~Brown \emph{et~al.}, ``Emu video: Factorizing text-to-video generation by explicit image conditioning,'' \emph{arXiv: 2311.10709}, 2023.

\bibitem{kondratyuk2023videopoet}
D.~Kondratyuk, L.~Yu, X.~Gu, J.~Lezama, J.~Huang, R.~Hornung, H.~Adam, H.~Akbari, Y.~Alon, V.~Birodkar \emph{et~al.}, ``Videopoet: A large language model for zero-shot video generation,'' \emph{arXiv: 2312.14125}, 2023.

\bibitem{hotshot}


\bibitem{yang2019video}
L.~Yang, Y.~Fan, and N.~Xu, ``Video instance segmentation,'' in \emph{IEEE International Conference on Computer Vision}, 2019, pp. 5188--5197.

\bibitem{huang2019got}
L.~Huang, X.~Zhao, and K.~Huang, ``Got-10k: A large high-diversity benchmark for generic object tracking in the wild,'' \emph{IEEE Transactions on Pattern Analysis and Machine Intelligence}, vol.~43, no.~5, pp. 1562--1577, 2019.

\bibitem{deng2009imagenet}
J.~Deng, W.~Dong, R.~Socher, L.-J. Li, K.~Li, and L.~Fei-Fei, ``Imagenet: A large-scale hierarchical image database,'' in \emph{IEEE Conference on computer Vision and Pattern Recognition}, 2009, pp. 248--255.

\end{thebibliography}

\end{document}